%% file: parafac2.tex
\documentclass[sigconf,table]{acmart}
\usepackage{booktabs} 
\usepackage{csquotes,xspace,algorithm,algorithmic,enumitem,color,subfigure}
\hypersetup{
    colorlinks=true
}

\newcommand{\T}[1]{\ensuremath{\mathcal{#1}}} 
\newcommand{\M}[1]{\ensuremath{\mathbf{#1}}} 
\newcommand{\V}[1]{\ensuremath{\mathbf{#1}}} 

\setcopyright{rightsretained}
\acmDOI{10.475/123_4}
\acmISBN{123-4567-24-567/08/06}
\acmConference[KDD'17]{ACM conference}{August 2017}{Halifax - Nova Scotia, Canada} 
\acmYear{2017}
\copyrightyear{2016}
\acmPrice{15.00}

\newcommand{\mname}{\texttt{SPARTan}\xspace}

\newcommand{\hide}[1]{}

\begin{document}
\title{\mname: Scalable PARAFAC2 for Large \& Sparse Data}

\author{Ioakeim Perros}
\affiliation{%
  \institution{Georgia Institute of Technology}
}

\author{Evangelos E. Papalexakis}
\affiliation{%
  \institution{University of California, Riverside}
}

\author{Fei Wang}
\affiliation{%
  \institution{Weill Cornell Medicine}
}

\author{Richard Vuduc}
\affiliation{%
  \institution{Georgia Institute of Technology}
}

\author{Elizabeth~Searles, Michael~Thompson}
\affiliation{%
  \institution{Children's Healthcare Of Atlanta}
}

\author{Jimeng Sun}
\affiliation{%
  \institution{Georgia Institute of Technology}
}

\input{content/abstract-rv} 

%
%

\ccsdesc[500]{Information systems~Data mining}
\keywords{Sparse Tensor Factorization, PARAFAC2, Phenotyping, Unsupervised learning}

\maketitle
\input{content/intro}
\input{content/background}
\input{content/sparafac2}

\input{content/methodology}

\input{content/experiments}

\input{content/conclusion}

\bibliographystyle{plainnat}
\bibliography{sigproc} 

\end{document}

%% file: content/abstract-rv.tex
\begin{abstract}
  In exploratory tensor mining, a common problem is how to analyze a set of variables across a set of subjects whose observations do not align naturally. For example, when modeling medical features across a set of patients, the number and duration of treatments may vary widely in time, meaning there is no meaningful way to align their clinical records across time points for analysis purposes. To handle such data, the state-of-the-art tensor 
  model is the so-called PARAFAC2, 
  which yields interpretable and robust output and can naturally handle sparse data. However, its main limitation up to now has been the lack of efficient algorithms that can handle large-scale datasets.

In this work, we fill this gap by developing a scalable method to compute the PARAFAC2 decomposition of large and sparse datasets, called \mname. Our method exploits special structure within PARAFAC2, leading to a novel algorithmic reformulation that is both fast (in absolute time) and more memory-efficient than prior work. We evaluate \mname on both synthetic and real datasets, showing $22\times$ performance gains over the best previous implementation and also handling larger problem instances for which the baseline fails. Furthermore, we are able to apply \mname to the mining of temporally-evolving phenotypes on data taken from real and medically complex pediatric patients. The clinical meaningfulness of the phenotypes identified in this process, as well as their temporal evolution over time for several patients, have been endorsed by clinical experts. 
\end{abstract}

%% file: content/intro.tex
\section{Introduction}\label{sec:intro}


\begin{table*}
\centering
\small
\begin{tabular}{|c|c|c|c|c||c|c|c|c|}
\hline
\textbf{Target Rank} & \multicolumn{4}{c||}{\textbf{10}} & \multicolumn{4}{c|}{\textbf{40}} \\ \hline
\textbf{\#nnz(Millions)} & \textbf{63} & \textbf{125} & \textbf{250} & \textbf{500} & \textbf{63} & \textbf{125} & \textbf{250} & \textbf{500} \\
\hline
\mname & 7.4 & 8.9 & 11.5 & 15.4 & 14 & 18.4 & 61 & 114 \\
\hline
Sparse PARAFAC2 & 24.4 & 60.1 & 72.3 & 194.5 & 275.2 & 408.1 & \textbf{\color{red}OoM} & \textbf{\color{red}OoM} \\
\hline
\end{tabular}
\caption{\footnotesize Running time comparison: Time in minutes of one iteration for increasingly larger datasets (63m to 500m) and fixed target rank (two cases considered: $R=\{10, 40\}$). The mode sizes for the datasets constructed are: $1$Mil.~subjects, $5$K variables and a maximum of $100$ observations per subject. \textbf{\color{red}OoM} (Out of Memory) denotes that the execution failed due to the excessive amount of memory requested. Experiments are conducted on a server with 1TB of RAM.}
\label{table:synthetic}
\end{table*}

This paper concerns tensor-based analysis and mining of multi-modal data where observations are difficult or impossible to align naturally along one of its modes.
A concrete example of such data is electronic health records (EHR), our primary motivating application.
An EHR dataset contains longitudinal patient information, represented as an event sequence of multiple modalities such as diagnoses, medications, procedures, and lab results.
An important characteristic of such event sequences is that there is no simple way to align observations in time across patients.
For instance, different patients may have varying length records between the first admission and the most recent hospital discharge;
or, two patients whose records' have the same length may still not have a sensible chronological alignment as disease stages and patient progress vary.

For tensor methods, such data poses a significant challenge.
Consider the most popular tensor analysis method in data mining, the canonical polyadic (CP) decomposition (also known as PARAFAC or CANDECOMP)~\cite{hitchcock1927expression,harshman1970foundations,carroll1970analysis}.
A dataset with three modes might be stored as an $I \times J \times K$ tensor $\T{X}$, which CP then decomposes into a sum of multi-way outer (rank-one) products,
$\T{X}~\approx~\sum_{r=1}^R \V{u}_r \circ \V{v}_r \circ \V{w}_r$, where $\V{u}_r, \V{v}_r, \V{w}_r $ are column vectors of size $I, J, K$, respectively, that effectively represents latent data concepts.
Its popularity owes to its \textit{intuitive output structure} and \emph{uniqueness} property that makes the model reliable to interpret~\cite{kruskal1977three,sidiropoulos2000uniqueness,kolda2009tensor,papalexakis2016tensors,sidiropoulos2016tensor}, as well as the existence of scalable algorithms and software~\cite{bader2007efficient,chi2012tensors,TTB_Software}.
However, to make, say, the irregular time points in EHR one of the input tensor modes would require finding some way to align time.
This fact is an inherent limitation of applying the CP model: any preprocessing to aggregate across time may lose temporal patterns~\cite{ho2014limestone,ho2014marble,wang2015rubik}, while more sophisticated temporal feature extraction methods typically need continuous and sufficiently long temporal measures to work~\cite{sun2008two}.
Other proposed methods specific to healthcare applications may give some good results~\cite{zhou2014micro,wang2013framework,wang2014densitytransfer} but lack the \textit{uniqueness} guarantee;
thus, it becomes harder to reliably extract the actual latent concepts as an equivalent arbitrary rotation of them will provide the same fit.
All of these weaknesses apply in the EHR scenario outlined above.


\begin{figure}[!b]
\centering
\includegraphics[scale=0.15]{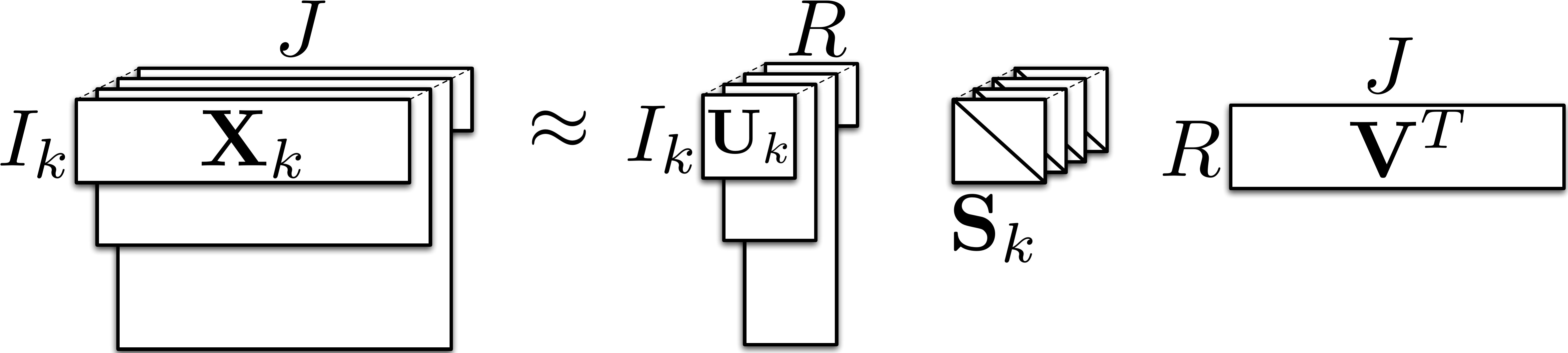}
\caption{\footnotesize Illustration of the PARAFAC2 model.}
\label{fig:parafac2}
\end{figure}

In fact, the type of data in the motivating example are quite general: 
consider that we have $K$ subjects, for which we record $J$ variables and we permit each $k$-th subject to have $I_k$ observations, which are not necessarily comparable among the different subjects.
For this type of data, Harshman proposed the \emph{PARAFAC2 model}~\cite{Hars1972b}.
It is a more flexible version of CP:
while CP applies the same factors across a collection of matrices, PARAFAC2 instead applies the same factor along one mode and allows the other factor matrix to vary~\cite{kolda2009tensor}.  
At the same time, it preserves the desirable properties of CP, such as uniqueness~\cite{harshman1996uniqueness,ten1996some,kiers1999parafac2,stegeman2015multi}.
As shown in Figure~\ref{fig:parafac2}, PARAFAC2 approximates each one of the input matrices as: $\M{X}_k \approx \M{U}_k ~ \M{S}_k ~ \M{V}^T$, where $\M{U}_k$ is of size $I_k \times R, \M{S}_k$ is a diagonal $R$-by-$R$, $\M{V}$ is of size $J \times R$ and $R$ is the target rank of the decomposition.  

Despite its applicability, the lack of efficient PARAFAC2 decomposition algorithms has been cited as a reason for its limited popularity~\cite{bro1997parafac,chew2007cross}.
Overall, PARAFAC2 has been mostly used for dense data (e.g.,~\cite{kiers1999parafac2}) or sparse data with a small number of subjects~\cite{chew2007cross}.
To our knowledge, no work has assessed PARAFAC2 for large-scale sparse data, as well as the challenges arising by doing so. 

In this paper, we propose \mname (abbreviated from Scalable PARafac Two) to fill this gap, with a focus on achieving scalability on large and sparse datasets.
Our methodological advance is a new algorithm for scaling up the core computational kernel arising in the PARAFAC2 fitting algorithm.
\mname achieves the best of both worlds in terms of speed and memory efficiency: \textit{a)} it is \textit{faster} than a highly-optimized baseline in all cases considered for both real (Figures~\ref{fig:all_scale_rank}, \ref{fig:choa_scale_all_ranks}, \ref{fig:movielens_scale_all_ranks}) and synthetic (Table~\ref{table:synthetic}) datasets, achieving up to 
$22\times$ performance gain; \textit{b)} at the same time, \mname is \textit{more scalable}, in that it can execute in reasonable time for large problem instances when the baseline fails due to excessive memory consumption (Table~\ref{table:synthetic}).

We summarize our contributions as:
\begin{itemize}[leftmargin=5.5mm]
\item \textbf{Scalable PARAFAC2 method:} We propose \mname, a scalable algorithm fitting the PARAFAC2 model on large and sparse data.
\item \textbf{Evaluation on various datasets:} We evaluate the scalability of our approach using datasets originating from two different application domains, namely a longitudinal EHR and a time-evolving movie ratings' dataset, which is also publicly available. Additionally, we perform synthetic data experiments.
\item \textbf{Real-world case study}: We performed a case study of applying \mname on
temporal phenotyping over medically complex pediatric patients in collaboration with Children's Healthcare of Atlanta (CHOA). The phenotypes and temporal trends discovered were endorsed by a clinical expert from CHOA. 
\end{itemize}
To promote reproducibility, our code will be open-sourced. For reviewing purposes, we provide the corresponding link~\footnote{\url{https://goo.gl/ovQkQr}}.

%% file: content/background.tex
\section{Background}\label{sec:back}
Next we describe the necessary terminology and operations regarding tensors. Then, we provide an overview of the CP model and relevant fitting algorithm. In Table~\ref{tab:notations}, we summarize the notations used throughout the paper.
\begin{table}
\footnotesize
\centering
\begin{tabular}{|c|c|} 
 \hline 
 \textbf{Symbol} & \textbf{Definition} \\ \hline \hline
$\T{X}, \M{X}, \V{x}, x$ & Tensor, matrix, vector, scalar \\
$\M{X}^{\dagger}$ & Moore-Penrose pseudoinverse \\
$\M{X}(:, i)$ & Spans the entire $i$-th column of $\M{X}$ (same for tensors) \\
$\M{X}(i, :)$ & Spans the entire $i$-th row of $\M{X}$ (same for tensors) \\
$diag(\V{x})$ & Diagonal matrix with vector $\V{x}$ on the diagonal \\
$diag(\M{X})$ & Extract diagonal of matrix $\M{X}$ \\
$\M{X}_k$ & shorthand for $\M{X}(:, :, k)$ (k-th frontal slice of tensor $\T{X}$)\\
$\{ \M{X}_k \}$ & the collection of $\M{X}_k$ matrices, for all valid $k$ \\
$\M{X}_{(n)}$ & mode-$n$ matricization of tensor $\T{X}$ \\
$\circ$ & Outer product \\
$\otimes$ & Kronecker product \\
$\odot$ & Khatri-Rao product \\
$*$ & Hadamard (element-wise) product \\
\hline
\end{tabular}
\caption{\footnotesize Notations used throughout the paper.}
\label{tab:notations}
\end{table}


\subsection{Tensors and Tensor Operations}
The \textit{order} of a tensor denotes the number of its dimensions, also known as ways or modes (e.g.,~matrices are $2$-order tensors).  A \textit{fiber} is a vector extracted from a tensor by fixing all modes but one. For example, a matrix column is a mode-$1$ fiber. 
A \textit{slice} is a matrix extracted from a tensor by fixing all modes but two. In particular, the $\M{X}(:, :, k)$ 
slices of a third-order tensor $\T{X}$ are called the frontal ones and we succinctly denote them as $\M{X}_k$~\cite{kolda2009tensor}. \textit{Matricization}, also called \emph{reshaping} or \emph{unfolding}, logically reorganizes tensors into other forms without changing the values themselves. The mode-$n$ matricization of a $N$-order tensor $\T{X} \in \mathbb{R}^{I_1 \times I_2 \times \dots \times I_N}$ is denoted by $\M{X}_{(n)} \in \mathbb{R}^{I_n \times I_1 I_2 \dots I_{n-1} I_{n+1} \dots I_N}$ and arranges the mode-$n$ fibers of the tensor as columns of the resulting matrix.

\subsection{CP Decomposition}\label{sec:back_cp}
The CP decomposition~\cite{hitchcock1927expression,harshman1970foundations,carroll1970analysis} of a third-order tensor $\T{X} \in \mathbb{R}^{I\times J \times K}$ is its approximation by a sum of three-way outer products: 
\begin{equation}\label{eq:cp_outer}
\T{X} \approx \sum_{r=1}^R \V{u}_r \circ \V{v}_r \circ \V{w}_r
\end{equation}where $\V{u}_r \in \mathbb{R}^I, \V{v}_r \in \mathbb{R}^J$ and $\V{w}_r \in \mathbb{R}^K$ are column vectors. 
If we assemble the column vectors $\V{u}_r, \V{v}_r, \V{w}_r$ as: $\M{U} = [\V{u}_1 ~ \V{u}_2 ~ \dots  ~ \V{u}_R] \in \mathbb{R}^{I\times R}, \M{V} = [\V{v}_1 ~ \V{v}_2 ~ \dots  ~ \V{v}_R] \in \mathbb{R}^{J\times R}, \M{W} = [\V{w}_1 ~ \V{w}_2 ~ \dots  ~ \V{w}_R] \in \mathbb{R}^{K\times R}$, 
then $\M{U}, \M{V}, \M{W}$ are called the \textit{factor matrices}. 
Interpretation of CP is very intuitive: we consider that the input tensor can be summarized as $R$ latent concepts. Then, for each $r$-th concept, the vectors $(\V{u}_r, \V{v}_r, \V{w}_r)$ are considered as soft-clustering membership indicators, for the corresponding I, J and K elements of each mode. An equivalent formulation of Relation~\eqref{eq:cp_outer} w.r.t. the frontal slices $\M{X}_k$ of the input tensor $\T{X}$ is~\cite{bro_thesis}:
\begin{equation}\label{eq:cp_slicewise}
\M{X}_k \approx \M{U} ~ \M{S}_k ~ \M{V}^T
\end{equation}where $k=1, 2, \dots, K$ and $\T{S} \in \mathbb{R}^{R \times R \times K}$ is an auxiliary tensor. Each frontal slice $\M{S}_k$ of $\T{S}$ contains the row vector $\M{W}(k, :)$ along its diagonal: $\M{S}_k = diag(\M{W}(k, :))$. Relation~\eqref{eq:cp_slicewise} provides another viewpoint of interpreting the CP model, through its correspondence to the Singular Value Decomposition (SVD): each slice $\M{X}_k$ is decomposed to a set of factor matrices $\M{U}, \M{V}$ (similar to the singular vectors) which are common for all the slices, and a diagonal middle matrix (similar to the singular values) which varies for each $k$-th slice. Note, however, that no orthogonality constraints are imposed on $\M{U}, \M{V}$ of the CP model, as in the SVD~\cite{chew2007cross}.

\noindent{\bf Uniqueness.} A fundamental property of CP is uniqueness~\cite{kruskal1977three,sidiropoulos2000uniqueness}. The issue with non-uniqueness can be exemplified via matrix factorization as follows~\cite{kolda2009tensor,papalexakis2016tensors}: If a matrix $\M{X}$ is approximated by the product of $\M{A} \M{B}^T$, then it can also be approximated with the same error by  $\M{A} \M{Q} \M{Q}^{-1} \M{B}^T = \M{\tilde{A}} \M{\tilde{B}^T}$, for any invertible $\M{Q}$. Thus, we can easily construct two completely different sets of rank-one factors that sum to the original matrix. Inevitably, this hurts interpretability, since we cannot know whether our solution is an arbitrarily rotated version of the actual latent factors.
In contrast to matrix factorization or Tucker decomposition~\cite{kolda2009tensor}, Kruskal~\cite{kruskal1977three} proved that CP is unique, under the condition: $k_{\M{U}} + k_{\M{V}} + k_{\M{W}} \geq 2R + 2$, where $k_{\M{U}}$ is the $k$-rank of $\M{U}$, defined as the maximum value $k$ such that any $k$ columns are linearly independent. The only exception is related to elementary indeterminacies of scaling and permutation of the component vectors~\cite{kolda2009tensor,sidiropoulos2016tensor}.
In sum, the CP decomposition is pursuing 
the true underlying latent information of the input tensor and provides reliable interpretation for unsupervised approaches.

\noindent{\bf Fitting the CP model.} Perhaps the most popular algorithm for fitting the CP model is the CP-Alternating Least Squares (CP-ALS)~\cite{carroll1970analysis,harshman1970foundations}, listed in Algorithm~\ref{alg:cp_als}. The main idea is to solve for one factor matrix at a time, by fixing the others. In that way, each subproblem is reduced to a linear least-squares problem. In case the input tensor contains non-negative values, a non-negative least-squares solver (e.g.,~\cite{bro1997fast}) can be used instead of an unconstrained one, to further improve the factors' interpretability~\cite{bro1997parafac}. 

Due to the ever increasing need for CP decompositions in data mining, the parallel CP-ALS for sparse tensors has been extensively studied in the recent literature for both single-node and distributed settings (e.g.,~\cite{bader2007efficient,kang2012gigatensor,choi2014dfacto,papalexakis2015p,smith2015splatt,cheng2016spals}). A pioneering work in addressing scalability issues for sparse tensors was provided by Bader and Kolda~\cite{bader2007efficient}~\footnote{The contributions of~\cite{bader2007efficient}, among others, are summarized as the Tensor Toolbox~\cite{TTB_Software}, which is widely acclaimed as the state-of-the-art package for single-node sparse tensor operations and algorithms.}. The authors identified and scaled up the algorithm's bottleneck, which is the materialization of the Matricized-Tensor-Times-Khatri-Rao-Product (MTTKRP). For example, in Algorithm~\ref{alg:cp_als}, the MTTKRP corresponds to the computation of $\M{X}_{(1)} (\M{W} \odot \M{V})$ when solving for $\M{U}$. For large and sparse tensors, a naive construction of the MTTKRP requires huge storage and computational cost and has to be avoided.

\begin{algorithm}
    \caption{\footnotesize CP-ALS}
    \label{alg:cp_als}
    \footnotesize
    \begin{algorithmic}[1]
    	\REQUIRE $\T{X} \in \mathbb{R}^{I \times J \times K}$ and target rank $R$
    	\ENSURE $\V{\lambda} \in \mathbb{R}^R, \M{U}\in \mathbb{R}^{I\times R}, \M{V}\in \mathbb{R}^{J\times R}, \M{W}\in \mathbb{R}^{K\times R}$
        \STATE Initialize $\M{V}, \M{W}$
        \WHILE{convergence criterion is not met}
        	\STATE $\M{U} \gets \M{X}_{(1)} (\M{W} \odot \M{V}) (\M{W}^T \M{W} * \M{V}^T \M{V})^{\dagger}$
            \STATE Normalize columns of $\M{U}$
            \STATE $\M{V} \gets \M{X}_{(2)} (\M{W} \odot \M{U}) (\M{W}^T \M{W} * \M{U}^T \M{U})^{\dagger}$
            \STATE Normalize columns of $\M{V}$
            \STATE $\M{W} \gets \M{X}_{(3)} (\M{V} \odot \M{U}) (\M{V}^T \M{V} * \M{U}^T \M{U})^{\dagger}$
            \STATE Normalize columns of $\M{W}$ and store norm in $\V{\lambda}$
        \ENDWHILE
    \end{algorithmic}
\end{algorithm}

%% file: content/sparafac2.tex
\section{PARAFAC2 Overview \& Challenges}
\label{sec:sparafac2}

\subsection{Model}
As we introduced in Section~\ref{sec:intro}, the PARAFAC2 model~\cite{Hars1972b} can successfully deal with an incomparable mode of each slice $\M{X}_k$~\cite{kiers1999parafac2}. It does so, by introducing a set of $\M{U}_k$ matrices replacing the $\M{U}$ matrix of the CP model in Relation~\eqref{eq:cp_slicewise}.  Thus, each slice $\M{X}_k$ is decomposed as shown in Figure~\ref{fig:parafac2}:
\begin{equation}\label{eq:paratwo_basic}
\M{X}_k \approx \M{U}_k ~ \M{S}_k ~ \M{V}^T
\end{equation} where $k=1, \dots, K, \M{U}_k \in \mathbb{R}^{I_k \times R}, \M{S}_k \in \mathbb{R}^{R \times R}$ is diagonal and $\M{V} \in \mathbb{R}^{J\times R}$. To preserve uniqueness, Harshman~\cite{Hars1972b} imposed the constraint that the cross product $\M{U}_k^T \M{U}_k$ is invariant regardless which subject $k$ is involved~\cite{kolda2009tensor,chew2007cross}. In that way, the CP model's invariance of the factor $\M{U}_k$ itself (or $\M{U}$ given its invariance to k), is relaxed~\cite{acar2009unsupervised}. 
For the above constraint to hold, each $\M{U}_k$ factor is decomposed as:
\begin{equation}
\M{U}_k = \M{Q}_k H
\end{equation} where $\M{Q}_k$ is of size $I_k \times R$ and has orthonormal columns, and $\M{H}$ is an $R \times R$ matrix, which does not vary by $k$~\cite{kolda2009tensor}. Then, the constraint that $\M{U}_k^T \M{U}_k$ is constant over $k$ is implicitly enforced, as follows: $\M{U}_k^T \M{U}_k = \M{H}^T \M{Q}_k^T \M{Q}_k \M{H} = \M{H}^T \M{H} = \M{\Phi}$. 

There have been several results regarding the uniqueness property of PARAFAC2~\cite{harshman1996uniqueness,ten1996some,kiers1999parafac2}. The most relevant towards the large-scale data scenario (i.e., the number of $K$ subjects can easily reach the order of hundreds of thousands), is that PARAFAC2 is unique for $K\geq 4$~\cite{stegeman2015multi}.



\subsection{Classical Algorithm for PARAFAC2}\label{sec:sparafac2_alg}
Below, we overview the classical algorithm for fitting PARAFAC2, proposed by \citet{kiers1999parafac2}. Their original algorithm expects dense data as input. Its objective function is as follows:
\begin{equation*}
\underset{ \{\M{U}_k\}, \{\M{S}_k\}, \M{V} }{\text{min}} \sum_{k=1}^K || \M{X}_k - \M{U}_k \M{S}_k \M{V}^T ||_F^2
\end{equation*}subject to: $\M{U}_k = \M{Q}_k \M{H},~\M{Q}_k^T \M{Q}_k = \M{I}$ and $\M{S}_k$ to be diagonal. Algorithm~\ref{alg:baseline} follows an Alternating Least Squares (ALS) approach, divided in two distinct steps: first (lines~\ref{alg:baseline_line_for_svd1}-\ref{alg:baseline_line_for_svd2}), the set of column-orthonormal matrices $\{ \M{Q}_k \}$ is computed by fixing $\M{H}, \M{V}, \{ \M{S}_k \}$. This step can be derived by examining the pursuit of each $\M{Q}_k$ as an individual Orthogonal Procrustes Problem~\cite{golub2013matrix} of the form: 
\begin{equation}\label{eq:procrustes_obj}
\underset{\M{Q}_k}{\text{min}} || \M{X}_k - \M{Q}_k \M{H} \M{S}_k \M{V}^T ||_F^2 
\end{equation} subject to $\M{Q}_k^T \M{Q}_k = \M{I}$. Given the SVD of $\M{H} \M{S}_k \M{V}^T \M{X}_k^T$ as $\M{P}_k \M{\Sigma}_k \M{Z}_k$, the minimum of objective~\eqref{eq:procrustes_obj} over column-orthonormal $\M{Q}_k$ is given by $\M{Q}_k = \M{Z}_k \M{P}_k^T$~\cite{kiers1999parafac2,golub2013matrix}.

Second, after solving for and fixing $\{ \M{Q}_k \}$, we find a solution for the rest of the factors as:
\begin{equation}\label{eq:cp_obj}
\underset{\M{H},\{\M{S}_k\}, \M{V}}{\text{min}} \sum_{k=1}^K || \M{Q}_k^T \M{X}_k -  \M{H} \M{S}_k \M{V}^T ||_F^2
\end{equation}where $\M{S}_k$ is diagonal. Note the equivalence of the above objective with the CP \enquote{slice-wise} formulation of Relation~\eqref{eq:cp_slicewise}. This equivalence implies that minimizing the objective~\eqref{eq:cp_obj} is achieved by executing the CP decomposition on a tensor $\T{Y}\in\mathbb{R}^{R \times J \times K}$ with frontal slices $\M{Y}_k = \M{Q}_k^T \M{X}_k$ (lines~\ref{alg:baseline_line_tensor}-\ref{alg:baseline_line_cp}). In order to avoid executing all the costly CP iterations, Kiers et al.~\cite{kiers1999parafac2} propose to run a single CP-ALS iteration, since this suffices to decrease the objective. 

The PARAFAC2 model can be extended so that non-negative constraints are imposed on $\{\M{S}_k\}, \M{V}$ factors~\cite{bro_thesis}. This is a property inherited by the CP-ALS iteration, where we can constrain the factors $\M{V}$ and $\M{W}$ to be non-negative, as discussed in Section~\ref{sec:back_cp}. Note that constraining the $\{\M{U}_k\}$ factors to be non-negative as well is not as simple, 
and a naive approach would violate the model properties~\cite{bro1999parafac2}.


\begin{algorithm}
    \caption{\footnotesize PARAFAC2-ALS~\cite{kiers1999parafac2}}
    \label{alg:baseline}
    \footnotesize
    \begin{algorithmic}[1]
    	\REQUIRE $\{\M{X}_k \in \mathbb{R}^{I_k \times J}\}$ for $k=1, \dots ,K$ and target rank $R$
    	\ENSURE $\{\M{U}_k \in \mathbb{R}^{I_k \times R}\}, \{\M{S}_k \in \mathbb{R}^{R\times R}\}$ for $k=1,\dots,K,~\M{V}\in\mathbb{R}^{J\times R}$
        \STATE Initialize $\M{H}\in\mathbb{R}^{R\times R}, \M{V},~\{\M{S}_k\}$ for $k=1, \dots, K$
        \WHILE{convergence criterion is not met}
        	\FOR{$k=1, \dots, K$}\label{alg:baseline_line_for_svd1}
            	\STATE $[\M{P}_k, \M{\Sigma}_k, \M{Z}_k] \gets $ truncated SVD of $\M{H} \M{S}_k \M{V}^T \M{X}_k^T$ at rank $R$
                \STATE $\M{Q}_k \gets \M{Z}_k \M{P}_k^T$ 
            \ENDFOR\label{alg:baseline_line_for_svd2}
           	\FOR{k=1, \dots, K}\label{alg:baseline_line_tensor}
				\STATE $\M{Y}_k \gets \M{Q}^T_k \M{X}_k$ \hspace{2.5cm} // \textit{construct slice $\M{Y}_k$ of tensor $\T{Y}$}
            \ENDFOR
            \STATE Run a single iteration of CP-ALS on $\T{Y}$ to compute $\M{H}, \M{V}, \M{W}$\label{alg:baseline_line_cp}
            \FOR{$k=1, \dots, K$}
            	\STATE $\M{S}_k \gets diag(\M{W}(k, :))$
            \ENDFOR
        \ENDWHILE
        \FOR{$k=1, \dots, K$}
        	\STATE $\M{U}_k \gets \M{Q}_k \M{H}$ \hspace{3cm}  // \textit{Assemble $\M{U}_k$}
        \ENDFOR
    \end{algorithmic}
\end{algorithm}

\subsection{Challenges of PARAFAC2 on sparse data}

Next, we summarize a set of crucial observations regarding the computational challenges, when executing Algorithm~\ref{alg:baseline} on large, sparse data:

\noindent{\bf Bottleneck of Algorithm~\ref{alg:baseline}.} Regarding the $1$st step (lines~\ref{alg:baseline_line_for_svd1}-\ref{alg:baseline_line_for_svd2}), in practice for sparse $\M{X}_k$, each one of the $K$ sub-problems scales as $\mathcal{O}(min(R I^2, R^2 I))$, due to the SVD involved~\cite{trefethen1997numerical}, where $I$ is an upper bound for $I_k$. Note that this computation can be trivially parallelized for all $K$ subjects. On the other hand, the $2$nd step (lines~\ref{alg:baseline_line_tensor}-\ref{alg:baseline_line_cp}) is dominated by the MTTKRP computation (which as we discussed in Section~\ref{sec:back} is the bottleneck of sparse CP-ALS). Thus, it scales as $3R~nnz(\T{Y})$~\cite{smith2015splatt}, using state-of-the-art sparse tensor libraries for single-node~\cite{bader2007efficient}. Given that none of the input matrices $\M{X}_k$ is completely zero, then: $3R~nnz(\T{Y}) \geq 3 K R^2$. As a result, this step becomes the bottleneck of Algorithm~\ref{alg:baseline} for large and sparse \enquote{irregular} tensors, since it cannot be parallelized w.r.t. the $K$ subjects, as trivially happens with the $1$st one.

\noindent{\bf Imbalance of mode sizes of $\T{Y}$.} The size of the intermediate tensor $\T{Y}$ formed is $R \times J \times K$. For large-scale data, we expect that $R << K, J$ since $R$ corresponds to the target rank of the overall PARAFAC2 decomposition. Note that this property of \enquote{size imbalance} may not hold for general large-scale sparse tensors, but it is crucial for optimizing PARAFAC2. 


\noindent{\bf Structured sparsity of $\{ \M{X}_k \}$.} First, we observe that even if the input slices $\{\M{X}_k \in \mathbb{R}^{I_k \times J}\}$ are very sparse, all their $I_k$ rows will contain at least one non-zero element. If this is not the case, we can simply filter the zero ones, without affecting the result. 
However, this does not hold for the $J$ columns of each $\M{X}_k$, which have to be aligned across all $K$ subjects. A direct consequence of that can be easily noticed in real datasets. For example, in the EHR \enquote{irregular} tensor of observations $\times$ variables $\times$ subjects, very few variables (e.g., diagnostic codes) are recorded for each subject. 
Driven by this observation, we are motivated to computationally exploit any column sparsity (i.e., cases where many columns will be completely zero) of each one of the input matrices $\M{X}_k$.



%% file: content/methodology.tex
\section{The \mname approach}\label{sec:method}
\subsection{Overview}\label{sec:method_overview}
Motivated by the challenges presented in Section~\ref{sec:sparafac2}, we propose a specialized version of the Matricized-Tensor-Times-Khatri-Rao-Product (MTTKRP) kernel, specifically targeting the intermediate tensor $\T{Y}  \in \mathbb{R}^{R \times J \times K}$ formed within the PARAFAC2-ALS algorithm. We first provide an overview of the properties that our approach exhibits:
\begin{itemize}[leftmargin=*]
\item It is {\em fully parallelizable w.r.t.~the $K$ subjects}. This property is crucial towards scaling up for large-scale \enquote{irregular} tensors.
\item It {\em exploits the structured sparsity} of the input frontal slices $\{\M{X}_k\}$. This is possible due to the observation that the $k$-th frontal slice $\M{Y}_k=\M{Q}_k^T \M{X}_k$ of $\T{Y}$ follows precisely the column sparsity pattern of $\M{X}_k$. In particular, if $c_k$ is the number of  columns of $\M{X}_k$ containing at least one non-zero element, then $\M{Y}_k$ will contain $R~c_k$ non-zero elements located in the positions of the non-zero columns of $X_k$. 
Exploiting structured sparsity is indispensable towards minimizing intermediate data and computations to the absolutely necessary ones.

\item As a by-product of the above, {\em \mname avoids unnecessary data re-organization} (tensor reshaping/permutations), since all operations are formulated w.r.t. the frontal slices $\M{Y}_k$ of tensor $\T{Y}$. In fact, our approach never forms the tensor $\T{Y}$ explicitly and directly utilizes the available collection  of matrices $\{ \M{Y}_k \}$ instead.
\end{itemize}

\subsection{Methodology}
In the following, we describe the design of our MTTKRP kernel for each one of the tensor modes.
We use the notation $\M{M}^{(i)}$ to denote the MTTKRP corresponding to the $i$-th tensor mode. Note that our factor matrices are: $\M{H} \in \mathbb{R}^{R \times R}, \M{V} \in \mathbb{R}^{J \times R}$ and $\M{W} \in \mathbb{R}^{K \times R}$ as in Line~\ref{alg:baseline_line_cp} of Algorithm~\ref{alg:baseline}.

\begin{figure}
\centering
\includegraphics[width=0.4\textwidth]{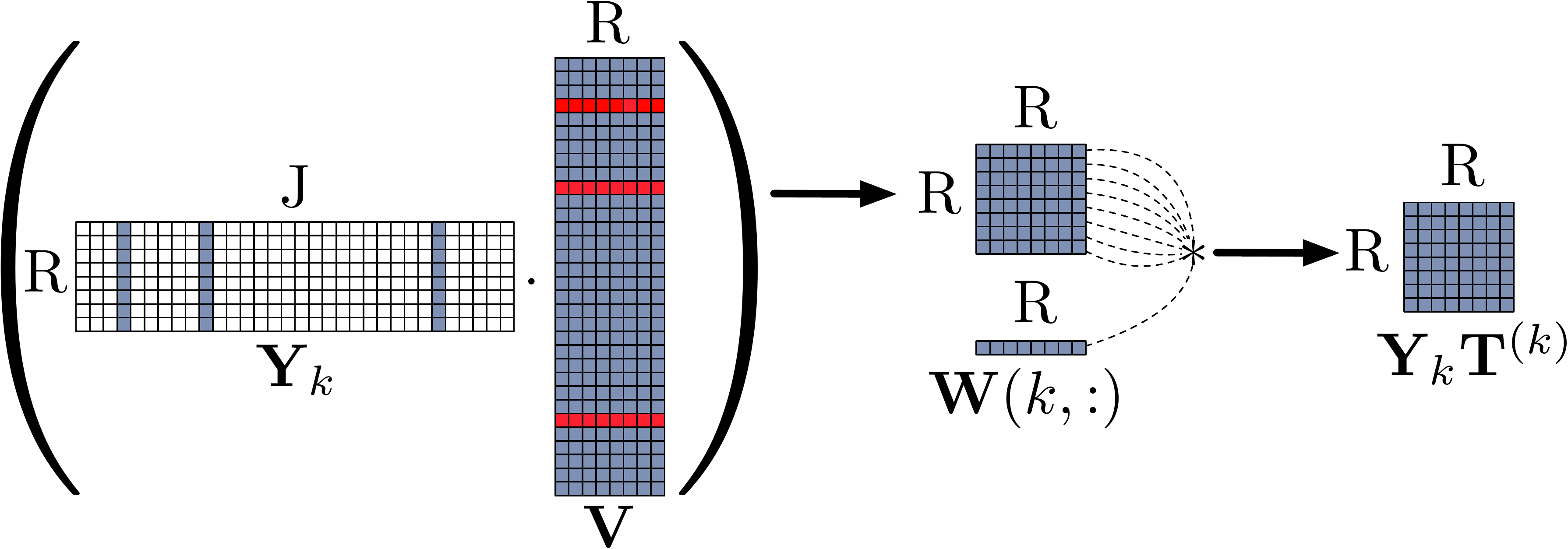}
\caption{\footnotesize \mname computations for the MTTKRP w.r.t.~the $1$st mode. For each $k$-th partial result of Equation~\eqref{eq:mttkrp1_2}, we use only the rows of $\M{V}$ factor matrix corresponding to the non-zero columns of $\M{Y}_k$. For each of the $R$ rows of the resulting matrix, we compute the Hadamard product with $\M{W}(k, :)$, which is the $k$-th row of the factor matrix $\M{W}$. The described computations fulfill all of the desirable properties presented in Section~\ref{sec:method_overview}. 
}
\label{fig:vis_mttkrp1}
\end{figure}

\noindent{\bf Mode-1 MTTKRP.} First, we re-visit the MTTKRP equation:
\begin{equation}\label{eq:mttkrp1}
\M{M}^{(1)} = \M{Y}_{(1)} \left( \M{W} \odot \M{V} \right),
\end{equation} where $\M{M}^{(1)} \in \mathbb{R}^{R \times R}, \M{Y}_{(1)} \in \mathbb{R}^{R \times KJ}$. 
 In order to attempt to parallelize the above computation w.r.t.~the $K$ subjects, we define the matrix $\M{T}^{(k)} \in \mathbb{R}^{J \times R}$ to denote the $k$-th vertical block of the Khatri-Rao Product $\M{W} \odot \M{V}\in \mathbb{R}^{KJ \times R}$:
{\small
\[
\M{W} \odot \M{V}=
\left[
\begin{array}{c}
\M{T}^{(1)} \\
\M{T}^{(2)} \\
\vdots \\
\M{T}^{(K)} 
\end{array}
\right]
\]
}We then remark that $\M{Y}_{(1)}$ (i.e., mode-$1$ matricization of $\T{Y}$) consists of an horizontal concatenation of the tensor's frontal slices $\M{Y}_k$. Thus, we exploit the fact that the matrix multiplication in Equation~\eqref{eq:mttkrp1} can be expressed as the sum of outer products or more generally, as a sum of block-by-block matrix multiplications:
\begin{equation}\label{eq:mttkrp1_2}
\M{M}^{(1)} = \sum_{k=1}^K \M{Y}_k ~ \M{T}^{(k)}
\end{equation}
Through Equation~\eqref{eq:mttkrp1_2}, the computation can be easily parallelized over $K$ independent sub-problems and then sum the partial results. This directly utilizes the frontal slices $\M{Y}_k$ without further tensor organization. However, it constructs the whole Khatri-Rao Product (in the form of blocks $\M{T}^{(k)}$). In order to avoid that, we first state an expression for each $i$-th row of $\M{T}^{(k)}$, which is a direct consequence of the Khatri-Rao Product definition:
\begin{equation}\label{eq:mttkrp1_1}
\M{T}^{(k)}(i, :) = \M{V}(i, :) * \M{W}(k, :),
\end{equation}
where $*$ stands for the Hadamard (element-wise) product. Then, we express the $j$-th row of each partial result of Equation~\eqref{eq:mttkrp1_2} as follows:
{\small
\begin{align} \label{eq:mttkrp1_3}
[\M{Y}_k ~ \M{T}^{(k)}]_{j, :} =&~ \M{Y}_k(j, :) ~ \M{T}^{(k)} \nonumber \\
\stackrel{Eq.\eqref{eq:mttkrp1_1}}{=}&~\sum_i \M{Y}_k(j, i) * \left( \M{V}(i,:) * \M{W}(k, :) \right) \nonumber \\
\stackrel{(a)}{=}&~\left( \sum_i \M{Y}_k(j, i) *  \M{V}(i,:)  \right) * \M{W}(k, :) \nonumber \\
\stackrel{(b)}{=}&~ \left( \M{Y}_k(j, :) ~ \M{V} \right) * \M{W}(k, :),\end{align}}where $(a)$ stems from the associative property of the Hadamard product and the fact that $\M{W}(k, :)$ is independent of the summation and $(b)$ from the calculation of matrix multiplication as a sum of outer-products (in particular, we encounter the sub-case of vector-matrix product).

Equation~\eqref{eq:mttkrp1_3} suggests an efficient way to compute the partial results of Equation~\eqref{eq:mttkrp1_2}, which we illustrate in Figure~\ref{fig:vis_mttkrp1}. First, we compute the matrix product $\M{Y}_k\M{V}$ and for each row of the intermediate result of size $R \times R$, we compute the Hadamard product with $\M{W}(k, :)$. Note that, as we discussed in Section~\ref{sec:sparafac2}, $\M{Y}_k$ is expected to be column-sparse in practice, thus multiplying by $\M{V}$ 
 uses only those rows of $\M{V}$ corresponding to the non-zero columns of $\M{Y}_k$. Thus, we avoid the redundant and expensive computation of the full Khatri-Rao Product. 
Overall, the methodology described above enjoys all of the properties described in Section~\ref{sec:method_overview}.

\begin{figure}
\centering
\includegraphics[width=0.3\textwidth]{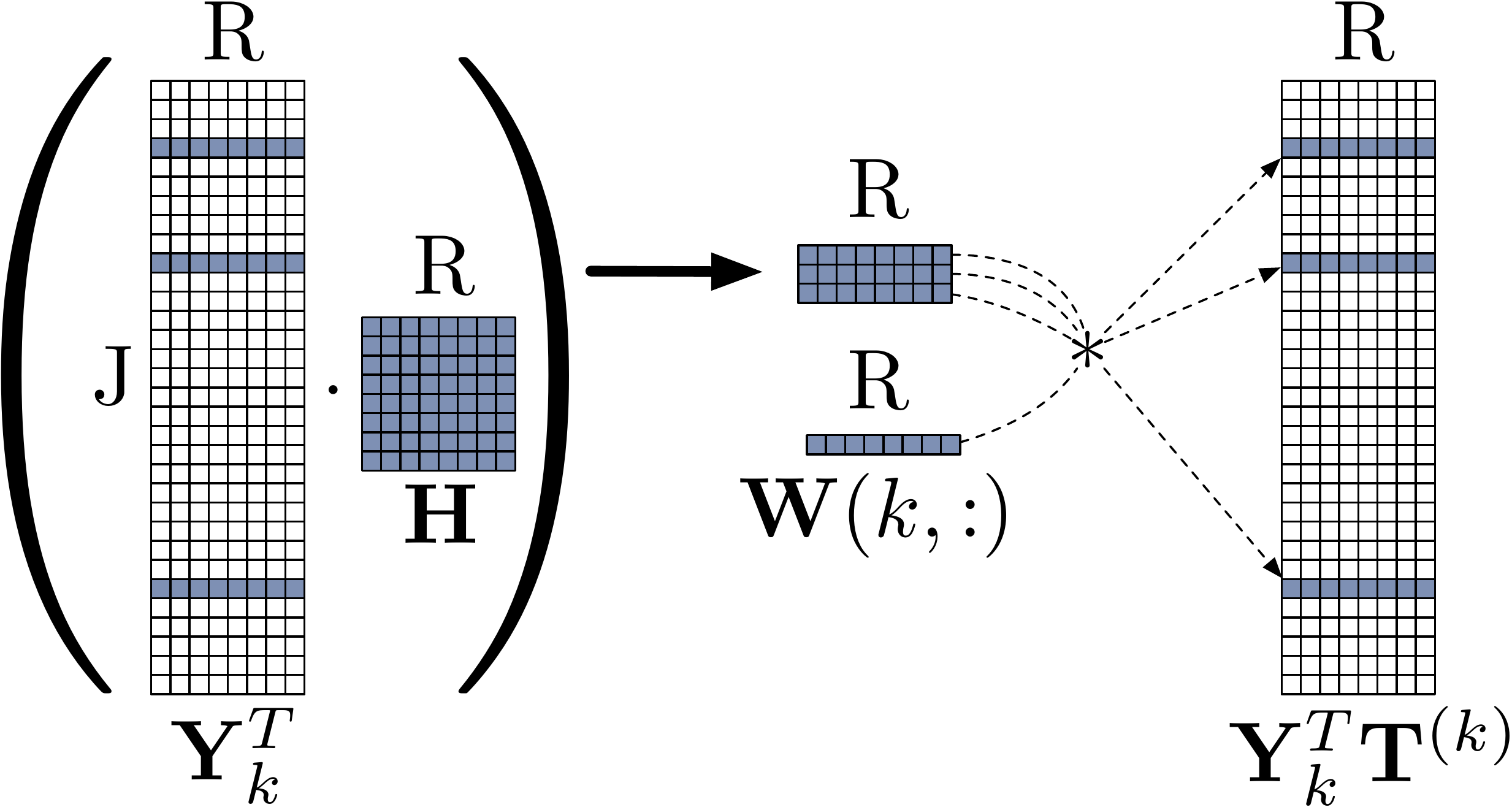}
\caption{\footnotesize \mname computations for the MTTKRP w.r.t.~the $2$nd mode. For each $k$-th partial result of Equation~\eqref{eq:mttkrp2_2}, we perform the vector-matrix multiplications for each non-zero row of $\M{Y}_k^T$. Then, for each intermediate vector, the Hadamard product with $\M{W}(k, :)$ is computed. Finally, we distribute the vectors to their corresponding positions in  $\M{Y}_k^T \M{T}^{(k)}$. As in the case w.r.t.~the $1$st mode, we limit computations to the necessary ones corresponding to the non-zero columns of $\M{Y}_k$ and all the properties presented in Section~\ref{sec:method_overview} are preserved.
}
\label{fig:vis_mttkrp2}
\end{figure}

\noindent{\bf Mode-2 MTTKRP.} The methodology followed for the Mode-$2$ case is similar to the one described for the $1$st case. We state the corresponding MTTKRP equation:
\begin{equation}\label{eq:mttkrp2}
\M{M}^{(2)} = \M{Y}_{(2)} \left( \M{W} \odot \M{H} \right)
\end{equation}
where $\M{M}^{(2)} \in \mathbb{R}^{J \times R}, \M{Y}_{(2)} \in \mathbb{R}^{J \times RK}$. 
The main remark is that $\M{Y}_{(2)}$ consists of an horizontal concatenation of the \textit{transposed} frontal slices $\{\M{Y}_k \}$ of the intermediate tensor $\T{Y}$. Thus, if we denote as $\M{T}^{(k)}$ the $k$-th vertical block of the Khatri-Rao Product $\M{W} \odot \M{H}$, we can formulate the problem as:
{\small
\begin{equation}\label{eq:mttkrp2_2}
\M{M}^{(2)} = \sum_{k=1}^K \M{Y}_k^T ~ \M{T}^{(k)}
 \end{equation}}
Given the above, it is easy to extend Equation~\eqref{eq:mttkrp1_3} for this case, so as to compute a single row of each partial result of Equation~\eqref{eq:mttkrp2_2}:
\begin{equation} \label{eq:mttkrp2_3}
[\M{Y}_k^T ~ \M{T}^{(k)}]_{j, :} = \left( \M{Y}_k(:, j)^T ~ \M{H} \right) * \M{W}(k, :)
\end{equation}
The corresponding operations are illustrated in Figure~\ref{fig:vis_mttkrp2}. A crucial remark is that we can focus on computing the relevant intermediate results only for the non-zero rows of $\M{Y_k}^T$, since the rest of the rows of the result $\M{Y_k}^T \M{T}^{(k)}$ will be zero. In sum, we again avoid redundant computations of the full Khatri-Rao Product and fulfill all of the properties described in Section~\ref{sec:method_overview}.

\begin{figure}
\centering
\includegraphics[width=0.4\textwidth]{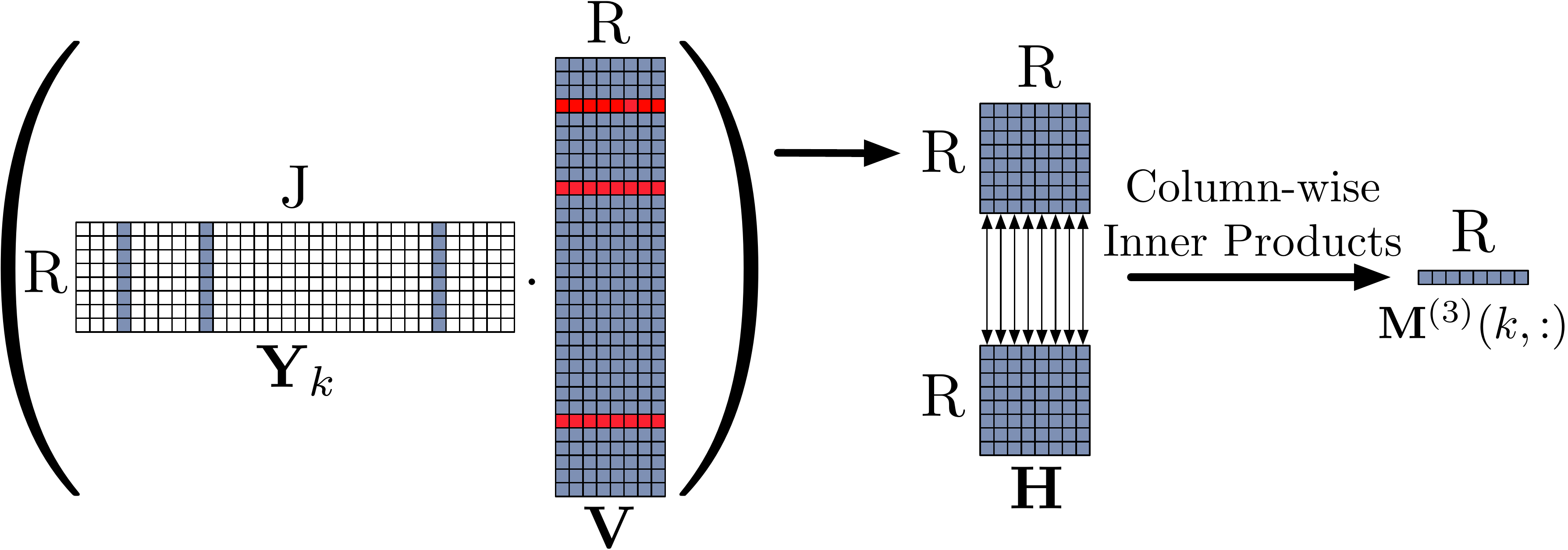}
\caption{\footnotesize \mname computations for the MTTKRP w.r.t.~the $3$rd mode. We compute each row of the result $\M{M}^{(3)}(k, :)$ independently of others, enabling parallelization w.r.t.~the $K$ subjects. As in mode-$1$, mode-$2$ cases, we exploit the column sparsity of $\M{Y}_k$. In this case, we also leverage that $\M{H}$ is a small $R$-by-$R$ matrix in practice (due to the \enquote{size imbalance} of the intermediate tensor $\T{Y}$). Thus, it is efficient to delay any computations on $\M{H}$ until the $R$-by-$R$ product of $\M{Y}_k \M{V}$ is formed, and then take column-wise inner products between those two matrices. The described operations fulfill all the properties outlined in Section~\ref{sec:method_overview}.}
\label{fig:vis_mttkrp3}
\end{figure}

\noindent {\bf Mode-3 MTTKRP.} First, we state the equation regarding the Mode-$3$ case:
\begin{equation}\label{eq:mttkrp3}
\M{M}^{(3)} = \M{Y}_{(3)} \left( \M{V} \odot \M{H} \right)
\end{equation}
where $\M{M}^{(3)} \in \mathbb{R}^{K \times R}$ and $\M{Y}_{(3)} \in \mathbb{R}^{K \times JR}$. Note that in this case, we are pursuing the MTTKRP of the mode corresponding to the $K$ subjects. Thus, an entirely different approach than the Mode-$1$, Mode-$2$ cases is needed so that we construct efficient independent sub-problems for each one of them. In particular, we need to design each $k$-th subproblem so that it computes the $k$-th row of $\M{M}^{(3)}$. In addition, we want to operate only on $\{\M{Y}_k\}$ without forming and reshaping the tensor $\T{Y}$, as well as to exploit the frontal slices' sparsity.
To tackle the challenges above, we leverage the fact that~\cite{choi2014dfacto}:
{\small
\begin{equation}\label{eq:mttkrp3_colwise}
\M{M}^{(3)}(:, r) =
\left[
\begin{array}{c}
{\M{H}(:, r)}^T~\M{Y}_1~\M{V}(:, r) \\
\vdots \\
{\M{H}(:, r)}^T~\M{Y}_K~\M{V}(:, r)
\end{array}
\right]
\end{equation}}Then, we remark that in order to retrieve a certain element of the matrix $\M{M}^{(3)}$, we have:
\begin{align*}
\M{M}^{(3)}(k, r) =&~{\M{H}(:, r)}^T ~ \M{Y}_k ~ \M{V}(:, r) \\
=&~{\M{H}(:, r)}^T ~ [\M{Y}_k \M{V}](:, r)
\end{align*}The last line above reflects the inner product between the corresponding $r$-th columns of $\M{H}$ and $[\M{Y}_k ~ \M{V}]$, respectively.
Thus, in order to retrieve a row $\M{M}(k, :)$, we can simply operate as:
\begin{equation}\label{eq:mttkrp3_row_wise}
\M{M}^{(3)}(k, :) = dot\left(\M{H}, \M{Y}_k\M{V}\right)
\end{equation}
where the $dot()$ function extracts the inner product of the corresponding columns of its two matrix arguments. We illustrate this operation in Figure~\ref{fig:vis_mttkrp3}. 
Since $\M{H}$ is a small $R$-by-$R$ matrix (due to the tensor's \enquote{size imbalance}), it is very efficient to delay any computations on $\M{H}$ until the $R$-by-$R$ intermediate matrix is formed as a product of $\M{Y}_k~\M{V}$. Then, we simply take the column-wise inner products between those two $R$-by-$R$ matrices. In that way, all the desirable properties we mentioned in Section~\ref{sec:method_overview} are also fulfilled.

In Algorithm~\ref{alg:spartan}, we list the pseudocode corresponding to the methodology proposed. Note that in lines~\ref{alg:spartan_1},\ref{alg:spartan_2}, 
we can accumulate over the partial results in parallel, since the summation is independent of the iteration order.

\begin{algorithm}
    \caption{\small MTTKRP for \mname}
    \label{alg:spartan}
    \small
    \begin{algorithmic}[1]
    	\REQUIRE $\{\M{Y}_k \in \mathbb{R}^{R \times J}\}$ for $k=1, \dots ,K$, $\M{H} \in \mathbb{R}^{R \times R}, \M{V} \in \mathbb{R}^{J \times R}, \M{W} \in \mathbb{R}^{K \times R}$, the target rank $R$ and the mode $n$ for which we are computing the MTTKRP
    	\ENSURE $\M{M}^{(n)}$
        \STATE Initialize $\M{M}^{(n)}$ with zeros
        \IF{$n==1$}
        	\FOR{$k=1, \dots, K$}
            	\STATE $temp \gets \M{Y}_k V$
                \FOR{$r=1, \dots, R$}
                	\STATE $temp(r, :) \gets temp(r, :) * \M{W}(k, :)$
                \ENDFOR
                \STATE $\M{M}^{(1)} \gets \M{M}^{(1)} + temp$ \hspace{0.5cm} // \textit{sum in parallel $\forall k=1, \dots, K$} \label{alg:spartan_1}
            \ENDFOR
        \ELSIF{$n==2$}
            \FOR{$k=1, \dots, K$}
            	\STATE Initialize $temp\in\mathbb{R}^{J\times R}$ with zeros
                \FOR{each $j$-th non-zero column of $\M{Y}_k$}
                	\STATE $temp(j, :) \gets \left( \M{Y}_k(:, j)^T \M{H} \right) * \M{W}(k, :)$
                \ENDFOR
                \STATE $\M{M}^{(2)} \gets \M{M}^{(2)} + temp$ \hspace{0.5cm} // \textit{sum in parallel $\forall k=1, \dots, K$} \label{alg:spartan_2}
            \ENDFOR
        \ELSIF{$n==3$} 
            \FOR{$k=1, \dots, K$}
            	\STATE $\M{M}^{(3)}(k, :) \gets dot\left(\M{H}, \M{Y}_k\M{V}\right)$ \hspace{0.5cm} // \textit{in parallel $\forall k=1, \dots, K$}
            \ENDFOR
        \ENDIF
    \end{algorithmic}
\end{algorithm}

%% file: content/experiments.tex
\section{Experiments}\label{sec:exp}
\subsection{Setup}\label{sec:exp_setup}
\begin{table}
\centering
\footnotesize
\begin{tabular}{|c|c|c|c|c|}
\hline 
Dataset & $K$ & $J$ & $max(I_k)$ & \#nnz \\ \hline \hline 
CHOA & 464,900 & 1,328 & 166 & 12.3 Mil. \\ \hline
MovieLens & 25,249 & 26,096 & 19 & 8.9 Mil. \\ \hline
\end{tabular}
\caption{\footnotesize Summary statistics for the real datasets of our experiments. $K$ is the number of subjects, $J$ is the number of variables, $I_k$ is the number of observations for the $k$-th subject and \#nnz corresponds to the total number of non-zeros.}
\label{table:data_stats}
\end{table}

\noindent{\bf Real Data Description.} Table~\ref{table:data_stats} provides summary statistics regarding the real datasets used. 

The \textbf{CHOA} (Children Healthcare of Atlanta) dataset corresponds to EHRs of pediatric patients with at least $2$ hospital visits. For each patient, we utilize the diagnostic codes and medication categories from their records, as well as the provided age of the patient (in days) at the visit time. The available International Classification of Diseases (ICD9)~\cite{slee1978international} codes are summarized to Clinical Classification Software (CCS)~\cite{ccs} categories, which is a standard step in healthcare analysis improving interpretability and clinical meaningfulness. 
We aggregate the time mode by week and all the medical events over each week are considered a single observation. The resulting data are of 464,900 subjects by 1,328 features by maximum 166 observations with 12.3m non-zeros. 


\textbf{MovieLens} 20M is another real dataset we used, which is \textit{publicly available}~\footnote{\url{https://grouplens.org/datasets/movielens/}}. We are motivated to use this dataset, because of the importance of the evolution of user preferences over time, as highlighted in recent literature~\cite{lathia2010temporal}. For this dataset, we consider that each year of ratings corresponds to a certain observation; thus, for each user, we have a  year-by-movie matrix to describe her rating activity. We consider only the users having at least $2$ years of ratings.

\noindent{\bf Implementation details.} We used MatlabR2015b for our implementations, along with functionalities for sparse tensors from the Tensor Toolbox~\cite{TTB_Software} and the Non-Negative Least Squares (NNLS) approach~\cite{bro1997fast} from the N-way Toolbox~\cite{nway}~\footnote{We also accredit the dense PARAFAC2 \href{http://www.models.life.ku.dk/algorithms}{implementation} by Rasmus Bro, from where we have adapted many functionalities.}. In both the \mname and the baseline implementations, we adjust the CP-ALS iteration arising in the PARAFAC2-ALS, so that non-negative constraints are imposed on the $\{ \M{S}_k \}, \M{V}$ factors, as discussed in Section~\ref{sec:sparafac2_alg}.



\noindent \textbf{The baseline method} corresponds to the standard fitting algorithm for the PARAFAC2 model~\cite{kiers1999parafac2} adjusted for sparse tensors as in~\cite{chew2007cross}. We utilized the implementation from the most recent version of the Tensor Toolbox~\cite{TTB_Software} regarding both the manipulation of sparse tensors, as well as the CP-ALS iteration arising in the PARAFAC2-ALS. 

\noindent \textbf{Parallelism.} We exploit the capabilities of the Parallel Computing Toolbox of Matlab, by utilizing its parallel pool in both \mname and the baseline approach, whenever this is appropriate. 
Regarding the size of the parallel pool, the number of workers of all the experiments regarding a certain dataset is fixed. For the movie-rating dataset we used the default of $12$ workers. For the synthetic and the CHOA datasets, 
we increased the number of workers to $20$ because of the data size increase. 

\noindent{\bf Hardware.} We conducted our experiments on a server running Ubuntu 14.04 with 1TB of RAM and four Intel E5-4620 v4 CPU's with a maximum clock frequency of 2.10GHz. Each one of the processors contains $10$ cores, and each one of the cores can exploit $2$ threads with hyper-threading enabled.
 
\subsection{\mname is fast and memory-efficient}
\noindent {\bf Synthetic Data.} We assess the scalability of the approaches under comparison for sparse synthetic data. We considered a setup with $1,000,000$ subjects, $5,000$ variables and a maximum of $100$ observations for each subject. The number of observations $I_k$ for each subject is dependent on the number of rows of $\M{X}_k$ containing non-zero elements; thus, $I_k$ increases with the dataset density. Indicatively, the mean number of observations $I_k$ for the sparsest dataset created ($\approx 63$ mil.) is $46.9$ and for the densest ($\approx 500$ mil.) dataset, the mean $I_k$ is $99.3$. We randomly construct the factors of a rank-$40$ (which is the maximum target rank used in our experiments) PARAFAC2 model. Based on this model, we construct the input slices $\{\M{X}_k\}$, which we then sparsify uniformly at random, for each sparsity level. The density of the sparsification governs the number of non-zeros of the collection of input matrices.

We provide the results in Table~\ref{table:synthetic}. First, we remark that
\mname is \textit{both more scalable} and \textit{faster} than the baseline. In particular, the baseline approach fails to execute in the two largest problem instances for target rank $R=40$, due to out of memory problems, during the creation of the intermediate sparse tensor $\T{Y}$. Note that as we discussed in Section~\ref{sec:method}, \mname avoids the additional overhead of explicitly constructing a sparse tensor structure, since it only operates directly on the tensor's frontal slices $\{\M{Y}_k\}$. Regarding the baseline's memory issue, since the density of $\T{Y}$ may grow (e.g., $\approx10\%$ in the densest case), we also attempted to store the intermediate tensor $\T{Y}$ as a dense one. However, this also failed, since the memory requested for a dense tensor of size $40$-by-$5$K-by-$1$Mil.~exceeded the available RAM of our system ($1$TB). Overall, it is clear that the baseline approach cannot fully exploit the input sparsity. On the contrary, \mname properly executes for all the problem instances considered in a reasonable amount of time. In particular, for $R=40$, \mname is up to $22\times$ faster than the baseline. Even for a lower target rank of $R=10$, \mname achieves up to $13\times$ faster computation.

\begin{figure}
	\centering
	\subfigure[{\footnotesize CHOA dataset}]{\includegraphics[width=1.6in]{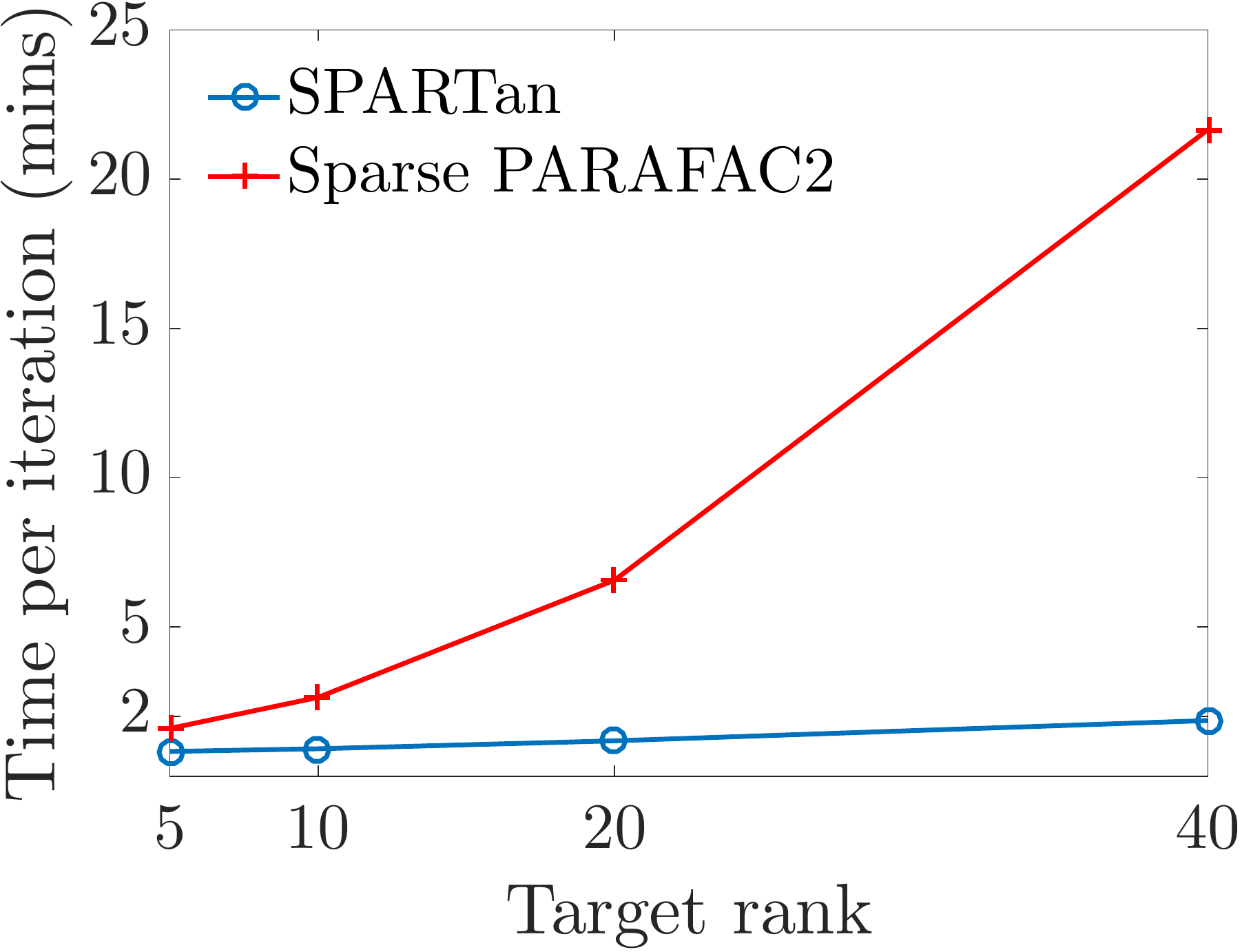}\label{fig:choa_scale}}
	\subfigure[{\footnotesize MovieLens dataset}]{\includegraphics[width=1.6in]{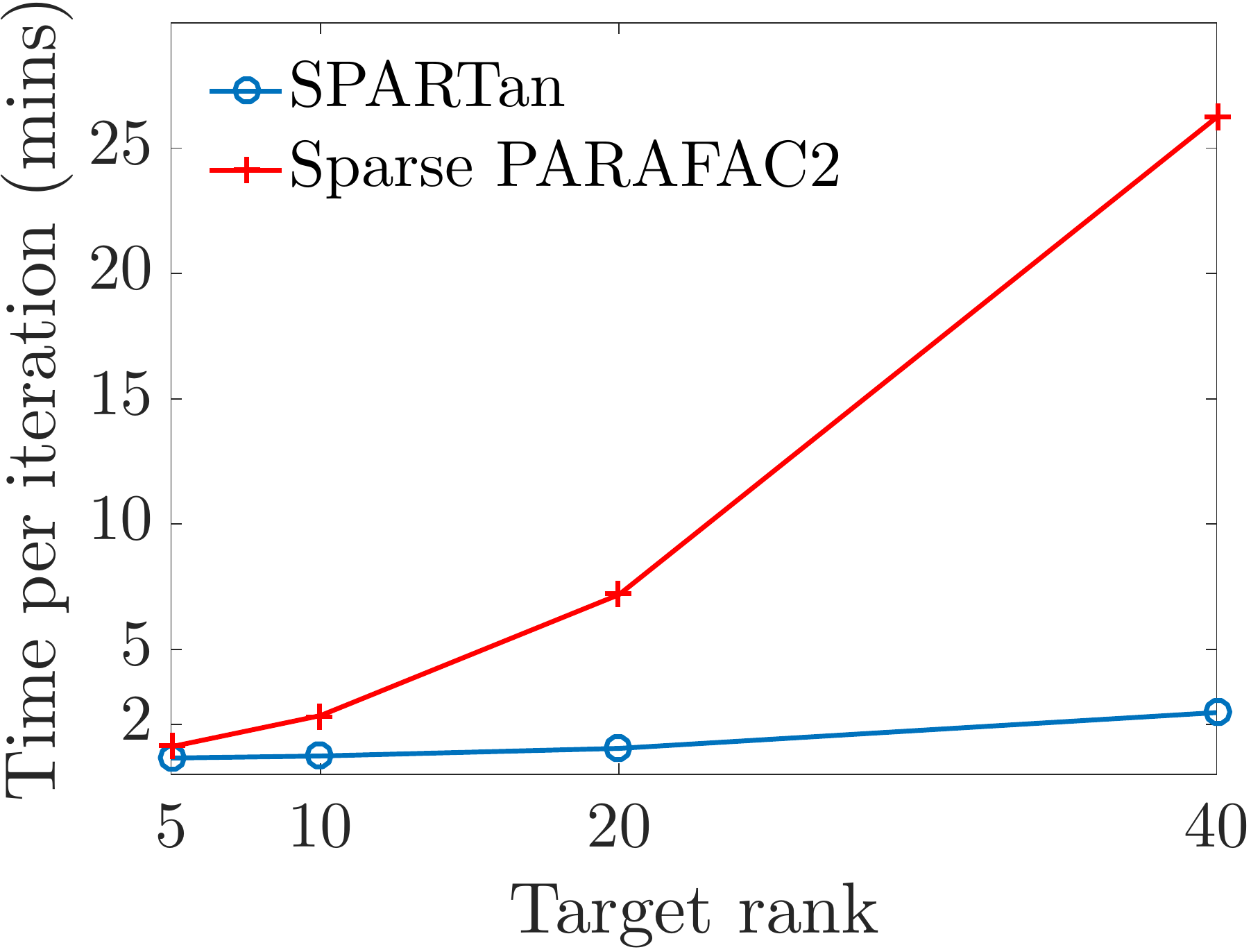}\label{fig:movielens_scale}}
	\caption{\footnotesize Time in minutes for one iteration (as an average over $10$) for varying target rank for both the real datasets used. \mname achieves up to $12\times$ and $11\times$ speedup over the baseline approach for the CHOA and the MovieLens datasets respectively.}
    \label{fig:all_scale_rank}
\end{figure}

\begin{figure}
	\centering
	\subfigure[{\footnotesize Target Rank $R=10$}]{\includegraphics[width=1.6in]{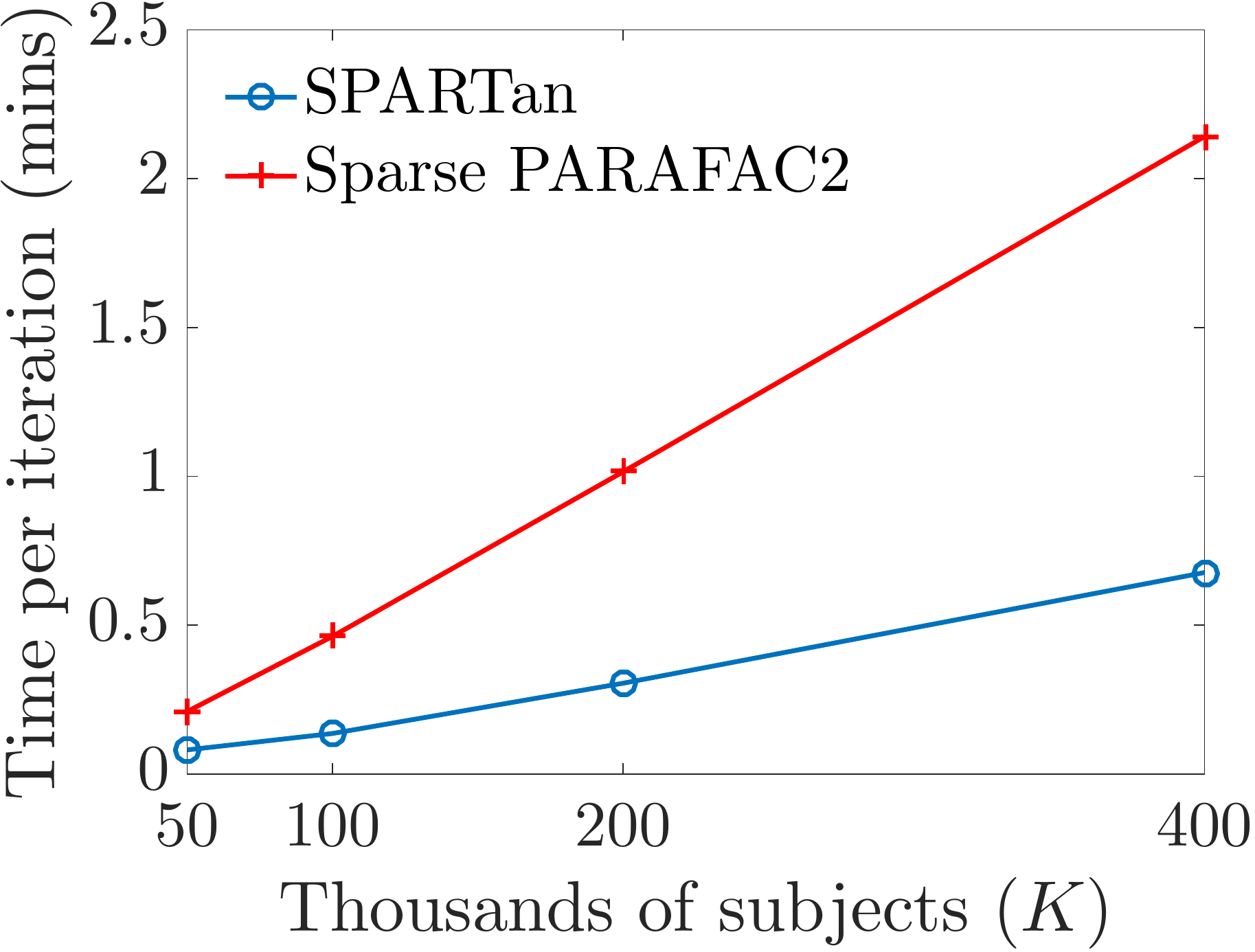}\label{fig:choa_scale_fixed_rank_10}}
	\subfigure[{\footnotesize Target Rank $R=40$}]{\includegraphics[width=1.6in]{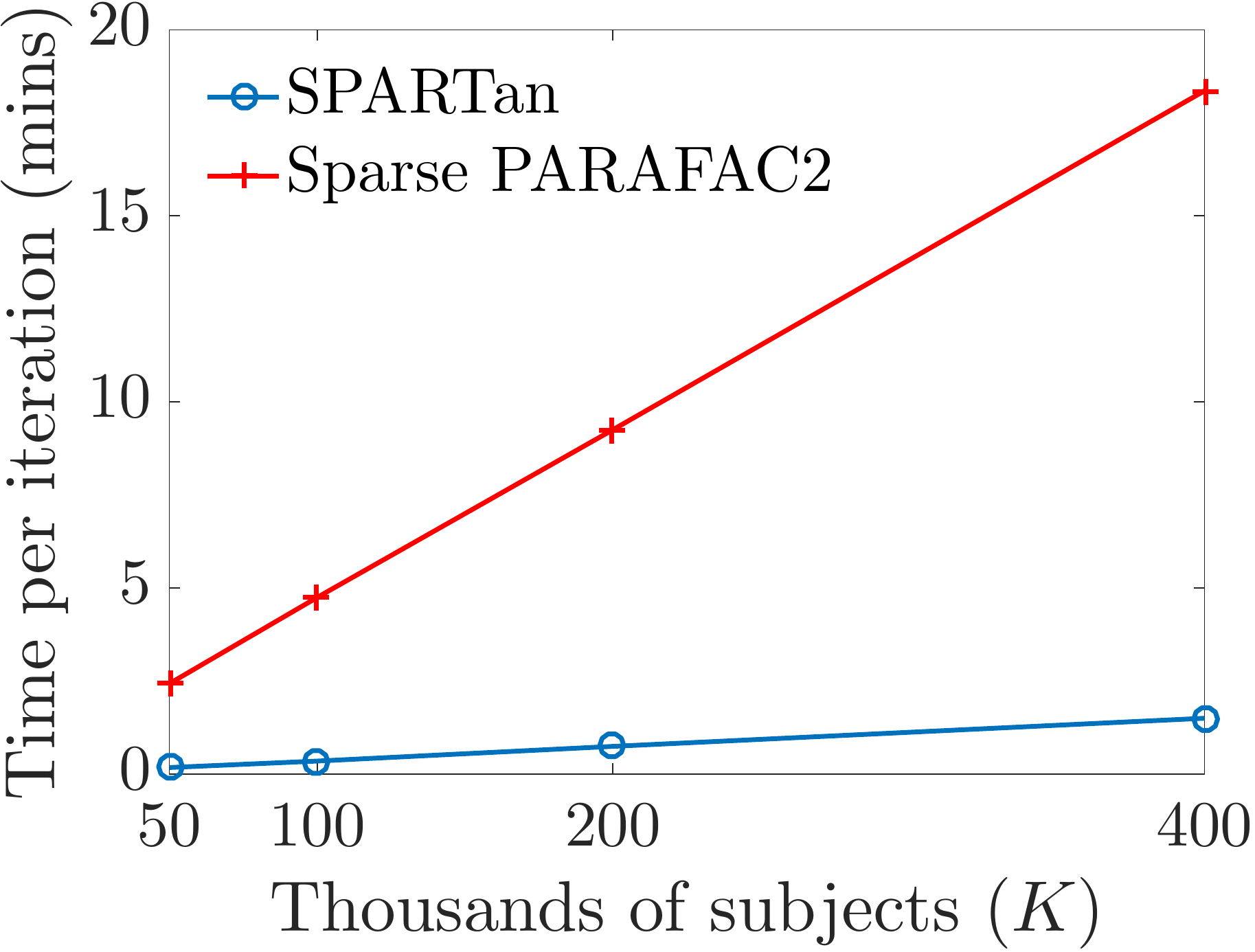}\label{fig:choa_scale_fixed_rank_40}}
	\caption{\footnotesize CHOA dataset: Time in minutes for one iteration (as an average over $10$) for varying number of subjects ($K$) included and fixed target rank (two cases considered: $R=\{10, 40\}$).}
    \label{fig:choa_scale_all_ranks}
\end{figure}

\begin{figure}
	\centering
	\subfigure[{\footnotesize Target Rank $R=10$}]{\includegraphics[width=1.6in]{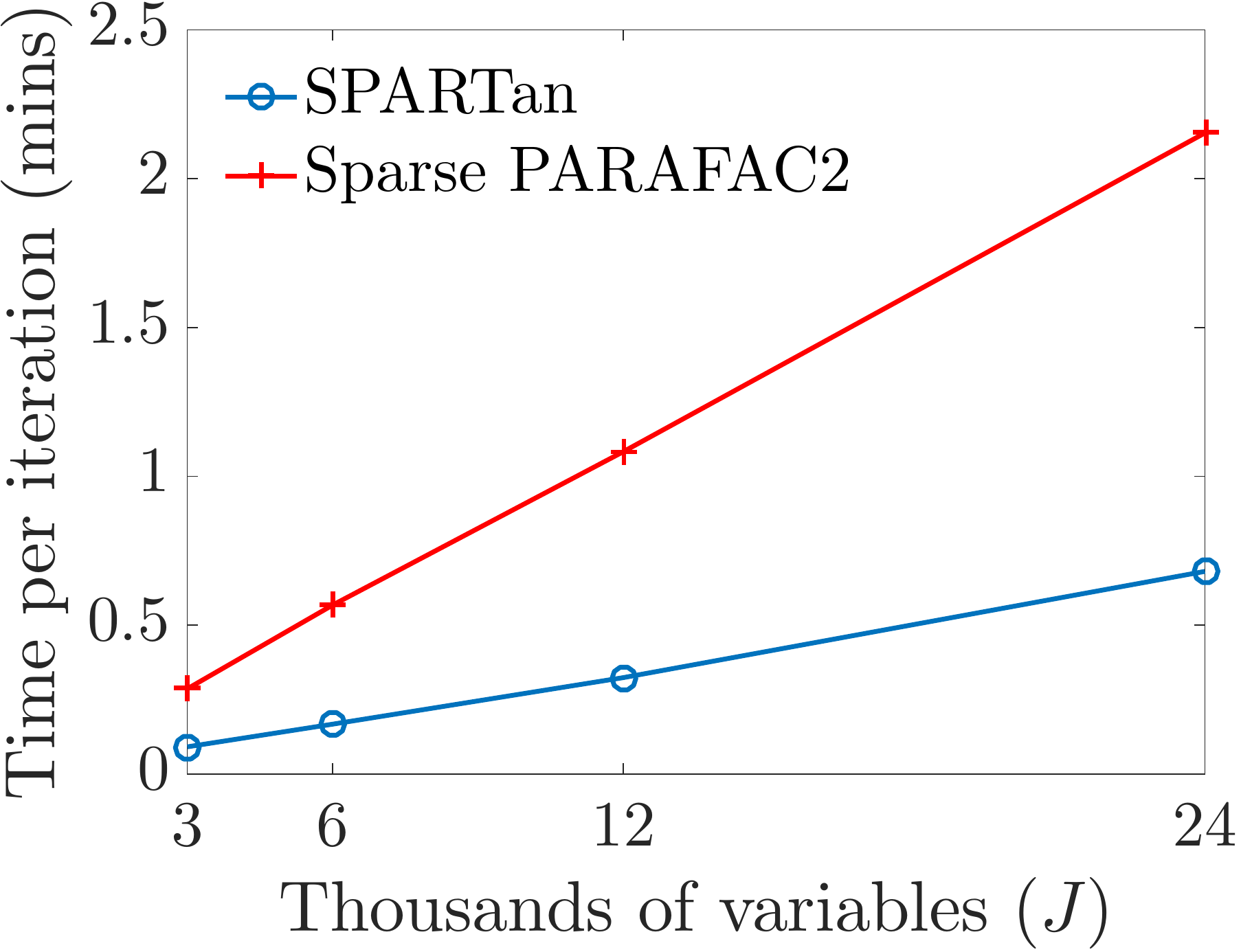}\label{fig:movielens_scale_fixed_rank_10}}
	\subfigure[{\footnotesize Target Rank $R=40$}]{\includegraphics[width=1.6in]{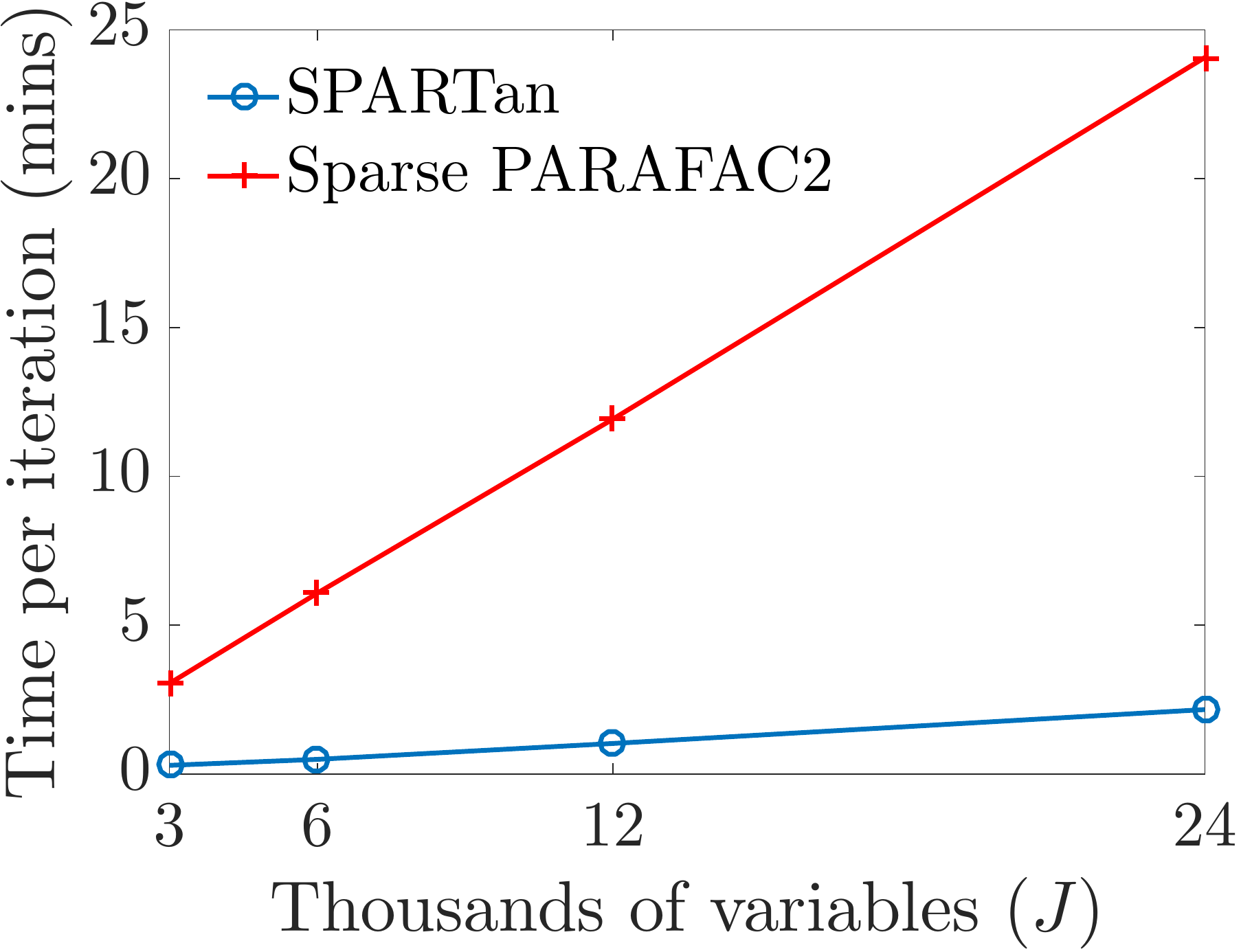}\label{fig:movielens_scale_fixed_rank_40}}
	\caption{\footnotesize MovieLens dataset: Time in minutes for one iteration (as an average over $10$) for varying number of variables ($J$) included and fixed target rank (two cases considered: $R=\{10, 40\}$).}
    \label{fig:movielens_scale_all_ranks}
\end{figure}


\noindent{\bf Real Data.}
We evaluate the scalability of the proposed \mname approach against the baseline method for the real datasets as well. In Figures~\ref{fig:all_scale_rank}, \ref{fig:choa_scale_all_ranks}, \ref{fig:movielens_scale_all_ranks}, we present the results of the corresponding experiments. First, we target the full datasets and vary the pursued target rank (Figure~\ref{fig:all_scale_rank}). Note that for both datasets considered, the time per iteration of the baseline approach increases dramatically as we increase the target rank. On the contrary, the time required by \mname increases only slightly. Overall, our approach achieves up to over an order of magnitude gain regarding the time required per epoch for both datasets.

We also evaluate the scalability of the methods under comparison as we vary the subjects and the variables considered. Since the CHOA dataset (Figure~\ref{fig:choa_scale_all_ranks}) contains more subjects than variables, we vary the number of subjects for this dataset for two fixed target ranks ($10, 40$). In both cases, \mname scales better than the baseline. As concerns the MovieLens dataset (Figure~\ref{fig:movielens_scale_all_ranks}), since it contains more variables than subjects, we examine the scalability w.r.t.~increasing subsets of variables considered. In this case as well, we remark the favorable scalability properties of \mname, rendering it practical to use for large and sparse \enquote{irregular} tensors.

\subsection{Phenotype discovery on CHOA EHR Data}\label{sec:exp_case_study}
\noindent{\bf Motivation.} Next we demonstrate the usefulness of PARAFAC2 towards temporal phenotyping of EHRs.  Phenotyping refers to the process of extracting meaningful patient clusters (i.e., phenotypes) out of raw, noisy Electronic Health Records~\cite{richesson2016clinical}. An open challenge in phenotyping is to capture temporal trends or patterns regarding the evolution of those phenotypes for each patient over time. Below, we illustrate how \mname can be used to successfully tackle this challenge. 


\noindent{\bf Model Interpretation:} We propose the following model interpretation towards the target challenge:
\begin{itemize}[leftmargin=5.5mm]
\item The common factor matrix  $\M{V}$ reflects the \textit{phenotypes' definition} and the non-zero values of each $r$-th column indicate the membership of the corresponding medical feature to the $r$-th phenotype.

\item The diagonal $\M{S}_k$ provides the \textit{importance membership indicators} of the $k$-th subject to each one of the $R$ phenotypes/clusters. Thus, we can sort the $R$ phenotypes based on the values of vector $diag(\M{S}_k)$ and identify the most relevant phenotypes for the $k$-th subject. 

\item Each $\M{U}_k$ factor matrix provides the \textit{temporal signature} of each patient: each $r$-th column of $\M{U}_k$ reflects the evolution of the expression of the $r$-th phenotype for all the $I_k$ weeks of her medical history. Note that since all $\M{X}_k, \M{S}_k, \M{V}$ matrices are non-negative, we only consider the non-negative elements of the temporal signatures in our interpretation.
\end{itemize}


\begin{figure}
\centering
\includegraphics[scale=0.075]{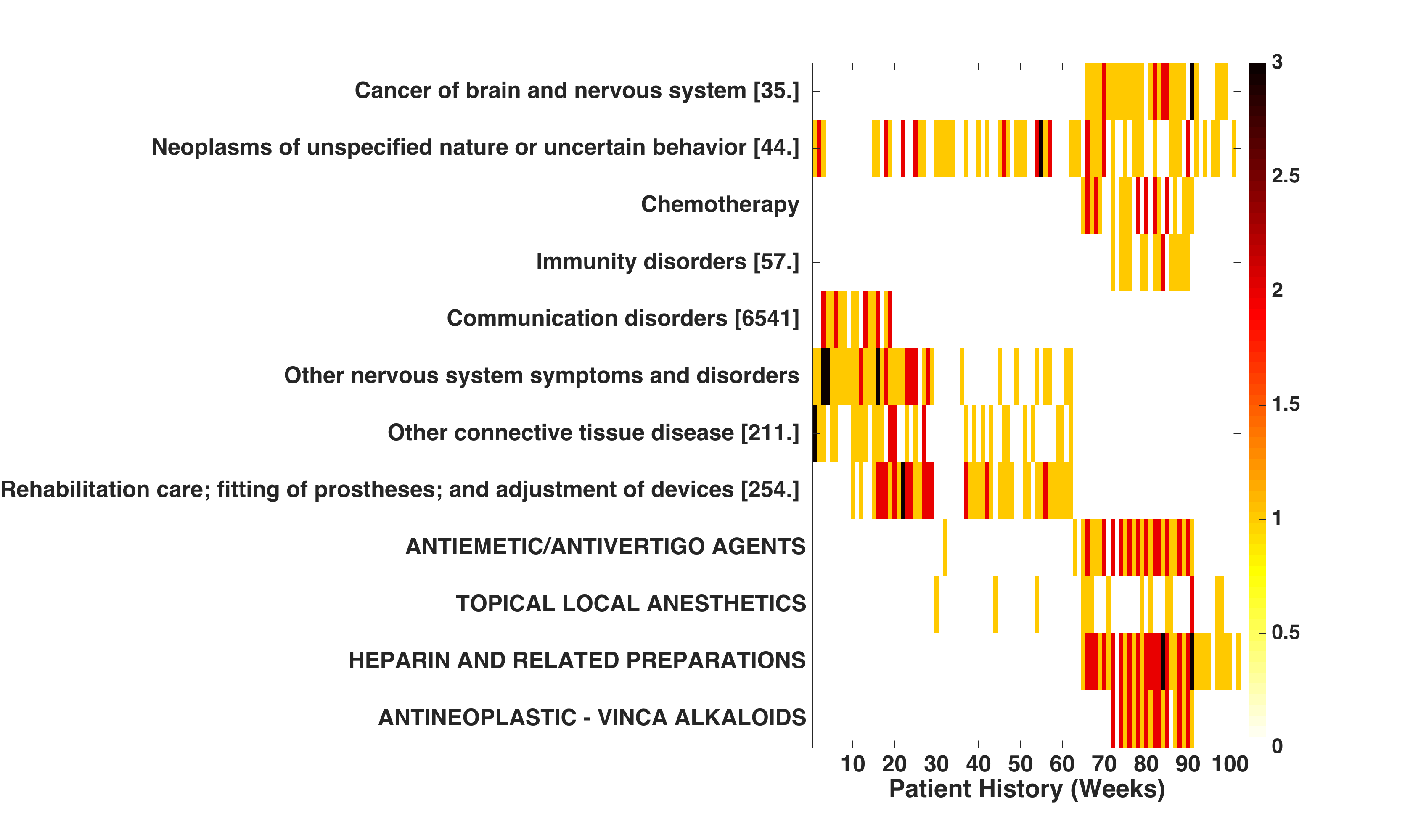}
\includegraphics[scale=0.35]{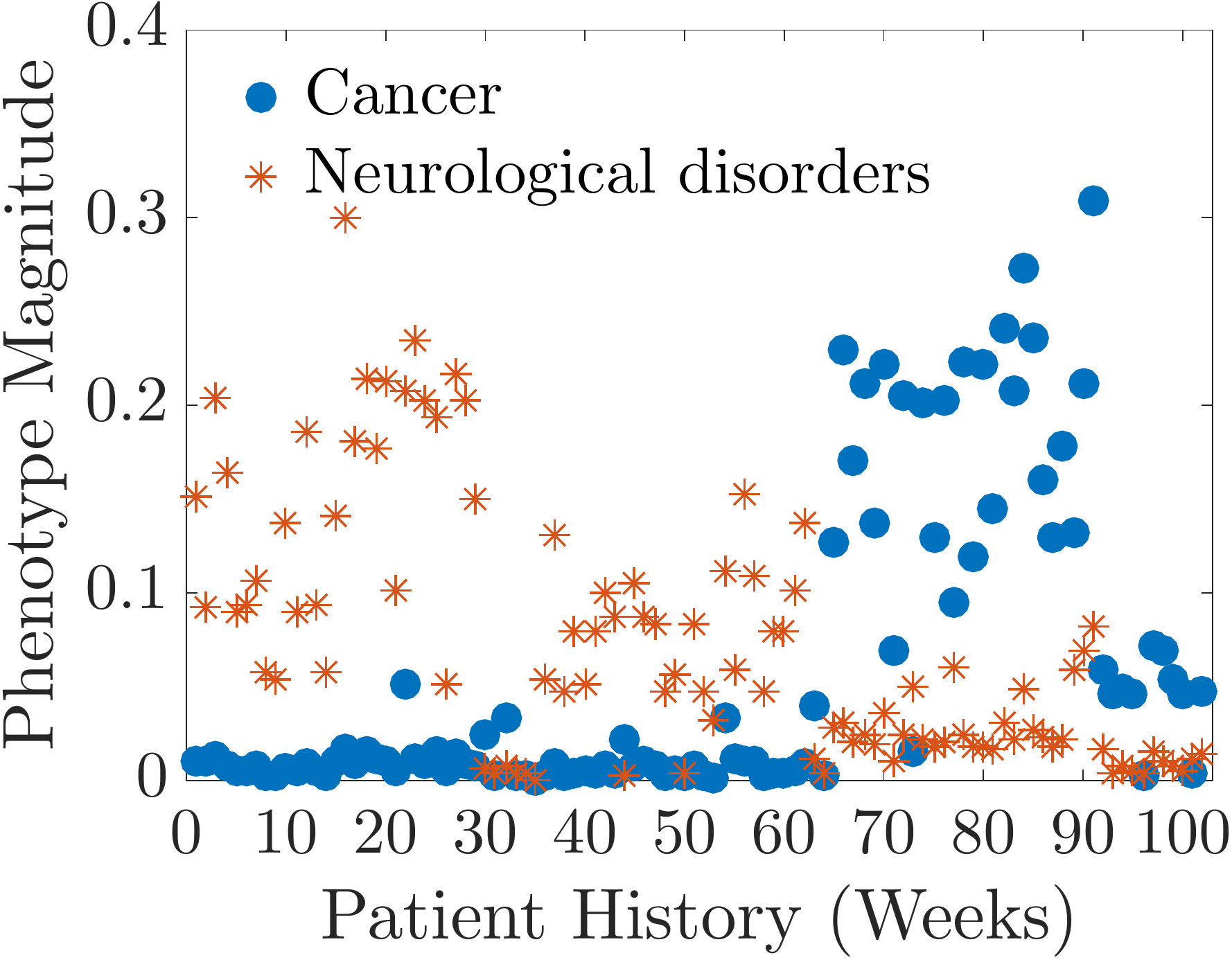}
\caption{\footnotesize \textit{Upper part:} Part of real EHR data of a Medically Complex Patient (MCP). For each week, it contains the occurrences of a diagnosis/medication in the patient's records. \textit{Lower part:} Temporal signature of the patient created by \mname. PARAFAC2 captures the stage where cancer treatment is initiated (week 65). At that point, indications of cancer treatment and diagnosis, such as cancer of brain, chemotherapy, heparin and antineoplastic drugs start to get recorded in the patient history. PARAFAC2 also captures the presence of neurological disorders during the first weeks of the patient history. The definition for each phenotype as produced by PARAFAC2 can be found in Table~\ref{tab:choa_pheno}.}
\label{fig:454_EHR}
\end{figure}
\input{results/table1.tex}

\noindent{\bf Temporal Phenotyping of Medically Complex Patients (MCPs)}
In order to illustrate the use of PARAFAC2 towards temporal phenotyping, we focus our analysis on a subset of pediatric patients from CHOA, which are classified by them as Medically Complex. These are the patients with high utilization, multiple specialty visits and high severity. Conceptually, those patients suffer from chronic and/or very severe conditions that are hard to treat. As a result, it becomes a very important challenge to accurately phenotype those patients, as well as provide a \textit{temporal signature} for each one of them, which summarizes their phenotypes' evolution.

The number of MCPs in the CHOA cohort is $8,044$, their diagnoses and medications sum up to $1,126$, and the mean number of weekly observations for those patients is $28$. We ran \mname for target rank $R=5$ and the $2$ phenotypes corresponding to our example patient in Figure~\ref{fig:454_EHR} are provided in Table~\ref{tab:choa_pheno} (\textit{phenotypes' definition} matrix). The rest of the phenotypes were also verified by the medical expert, but are omitted due to space limits. We provide the full phenotypes' list, as well as additional patient examples in the Supplementary Material~\footnote{\url{https://goo.gl/q7L91D}}. In Figure~\ref{fig:454_EHR}, we provide part of the real EHR, as well as the temporal signature produced by \mname, for a certain medically complex patient. Regarding the EHR, we visualize the subset of diagnoses and medications for which the sum of occurrences for the whole patient history is above a certain threshold (e.g., 5 occurrences). This step ensures that the visualized EHR will only contain the conditions exhibiting some form of temporal evolution. For the patient examples considered, we identify the top-$2$ relevant phenotypes through the importance membership indicator matrix $\M{S}_k$ as discussed above. For those top-$2$ phenotypes, we present the resulting \textit{temporal signature}, from which we easily detect intricate temporal trends of the phenotypes involved. Those trends were confirmed by the clinical expert as valuable towards fully understanding the phenotypic behavior of the MCPs.



%% file: results/table1.tex


\begin{table}[htbp]
  \centering
     \scriptsize
    \begin{tabular}{lc}
    \hline  \textbf{Cancer} & \multicolumn{1}{l}{\textbf{Weight}} \\ \hline 
    \textcolor[rgb]{ .753,  0,  0}{Chemotherapy } & 0.35 \\
    \textcolor[rgb]{ .753,  0,  0}{Leukemias [39.] } & 0.27 \\
    \textcolor[rgb]{ .753,  0,  0}{Immunity disorders [57.] } & 0.23 \\
    \textcolor[rgb]{ 0,  .439,  .753}{HEPARIN AND RELATED PREPARATIONS} & 0.6 \\
    \textcolor[rgb]{ 0,  .439,  .753}{ANTIEMETIC/ANTIVERTIGO AGENTS} & 0.34 \\
    \textcolor[rgb]{ 0,  .439,  .753}{SODIUM/SALINE PREPARATIONS} & 0.32 \\
    \textcolor[rgb]{ 0,  .439,  .753}{TOPICAL LOCAL ANESTHETICS} & 0.19 \\
    \textcolor[rgb]{ 0,  .439,  .753}{ANTIHISTAMINES - 1ST GENERATION} & 0.16 \\ \hline
   \textbf{Neurological System Disorders} & \multicolumn{1}{l}{\textbf{Weight}} \\ \hline
    \textcolor[rgb]{ .753,  0,  0}{Other nervous system symptoms and disorders } & 0.56 \\
    \textcolor[rgb]{ .753,  0,  0}{Rehabilitation care; fitting of prostheses; and adjustment of devices [254.] } & 0.5 \\
    \textcolor[rgb]{ .753,  0,  0}{Residual codes; unclassified; all E codes [259. and 260.] } & 0.46 \\
    \textcolor[rgb]{ .753,  0,  0}{Other connective tissue disease [211.] } & 0.33 \\
    \textcolor[rgb]{ .753,  0,  0}{Other and unspecified metabolic; nutritional; and endocrine disorders } & 0.18 \\
    \end{tabular}%
    \caption{\footnotesize Subset of phenotypes discovered by PARAFAC2 corresponding to Figure~\ref{fig:454_EHR}. The title annotation for each phenotype is provided by the medical expert. The red color corresponds to diagnoses and the blue color corresponds to medications.}
      \label{tab:choa_pheno}
\end{table}%

%% file: content/conclusion.tex
\section{Discussion \& Conclusions}\label{sec:concl}
PARAFAC2 has been the state-of-the-art model for mining \enquote{irregular} tensors, where the observations along one of its modes do not align naturally. However, it has been highly disregarded by practitioners, as compared to other tensor approaches. Bro~\cite{bro1997parafac} has summarized the reason for that as:
\begin{quote}
\textit{The PARAFAC2 model has not yet been used very extensively maybe because the implementations so far have been complicated and slow.}
\end{quote}
The methodology proposed in this paper renders this statement no longer true for large and sparse data. In particular, as tested over real and synthetic datasets, \mname is both fast and memory-efficient, achieving up to $22\times$ performance gains over the best previous implementation and also handling larger problem instances for which the baseline fails due to insufficient memory.

The key insight driving \mname's scalability is the pursuit and exploitation of special structure in the data involved in intermediate computations; prior art did not do so, instead treating those computations as a black-box.

The capability to run PARAFAC2 at larger scales is, in our view, an important enabling technology. As shown in our evaluations on EHR data, the clinically meaningful phenotypes and temporal trends identified by PARAFAC2 reflect the ease of the model's interpretation and its potential utility in other application domains.

Future directions include, but are not limited to: \textit{a)} development of PARAFAC2 algorithms for alternative models of computation, such as distributed clusters~\cite{kang2012gigatensor}, or supercomputing environments; \textit{b)} extension of the methodology proposed for higher-order \enquote{irregular} tensors with more than one mismatched mode.

Finally, to enable reproducibility and promote further popularization of the PARAFAC2 modeling within the area of data mining, we make our implementations \textit{publicly available}.
